%% file: main.tex
\documentclass[10pt,twocolumn,letterpaper]{article}

\usepackage{cvpr}
\usepackage{times}
\usepackage{epsfig}
\usepackage{graphicx}
\usepackage{amsmath}
\usepackage{amssymb}
\usepackage{nicefrac}
\usepackage{multirow}
\usepackage{tabularx, makecell}
\usepackage[justification=centering]{caption}
\usepackage[breaklinks=true,bookmarks=false]{hyperref}

\cvprfinalcopy % *** Uncomment this line for the final submission

\def\cvprPaperID{6349} % *** Enter the CVPR Paper ID here

\ifcvprfinal\fi
\begin{document}

%%%%%%%%% TITLE
\title{Feature Distillation: DNN-Oriented JPEG Compression Against Adversarial Examples}
\author{%\large
Zihao Liu$^1$,
Qi Liu$^1$, 
Tao Liu$^1$, 
Nuo Xu$^1$, 
Xue Lin$^2$,
Yanzhi Wang$^2$,
Wujie Wen$^1$ \\
$^1$ Flordia International University,
$^2$ Northeastern University\\
\{\tt\small zliu021,qliu020,tliu023,nxu003,wwen\}@fiu.edu,
\{\tt\small xue.lin, yanz.wang\}@northeastern.edu
}

% \author{Zihao Liu\\
% Flordia International University\\
% {\tt\small zliu021@fiu.edu}
% % For a paper whose authors are all at the same institution,
% % omit the following lines up until the closing ``}''.
% % Additional authors and addresses can be added with ``\and'',
% % just like the second author.
% % To save space, use either the email address or home page, not both
% \and
% Qi Liu\\
% Flordia International University\\
% {\tt\small qliu020@fiu.edu}
% \and
% Tao Liu\\
% Flordia International University\\
% {\tt\small tliu023@fiu.edu}
% \and
% Nuo Xu\\
% Flordia International University\\
% {\tt\small nxu003@fiu.edu}
% \and
% Nuo Xu\\
% Flordia International University\\
% {\tt\small nxu003@fiu.edu}
% }

\maketitle
%\thispagestyle{empty}

%%%%%%%%% ABSTRACT
\begin{abstract}
Image compression-based approaches for defending against the adversarial-example attacks, which threaten the safety use of deep neural networks (DNN), have been investigated recently. However, prior works mainly rely on directly tuning parameters like compression rate, to blindly reduce image features, thereby lacking guarantee on both defense efficiency (i.e. accuracy of polluted images) and classification accuracy of benign images, after applying defense methods. 
To overcome these limitations, we propose a JPEG-based defensive compression framework, namely ``feature distillation", to effectively rectify adversarial examples without impacting classification accuracy on benign data. 
Our framework significantly escalates the defense efficiency with marginal accuracy reduction using a two-step method: First, we maximize malicious features filtering of adversarial input perturbations by developing defensive quantization in frequency domain of JPEG compression or decompression, guided by a semi-analytical method; Second, we suppress the distortions of benign features to restore classification accuracy through a DNN-oriented quantization refine process. 
Our experimental results show that proposed ``feature distillation" can significantly surpass the latest input-transformation based mitigations such as Quilting and TV Minimization in three aspects,  including defense efficiency (improve classification accuracy from $\sim20\%$ to $\sim90\%$ on adversarial examples), accuracy of benign images after defense ($\le1\%$ accuracy degradation), and processing time per image ($\sim259\times$ Speedup).
Moreover, our solution can also provide the best defense efficiency ($\sim60\%$ accuracy) against the recent adaptive attack with least accuracy reduction ($\sim1\%$) on benign images when compared with other input-transformation based defense methods.
%Particularly, our method can also defense the latest adaptive adversarial attack BPDA (which is indefensible by most published methods) with significantly improved effectiveness.
\end{abstract}

%%%%%%%%% BODY TEXT
\input{1.intro}
\input{2.back}

\input{3.design}
\input{4.evaluation}

\newpage

{\small
\bibliographystyle{ieee}
\bibliography{egbib}
}

\end{document}

%% file: 1.intro.tex
\vspace{-10pt}
\section{Introduction}
\label{sec:intro}
%state-of-the-art input transformation methods on 3 aspects;
%our approach defeat BPDA, other;no published work can in rebuttal-review2

% Thanks to the recent machine learning model innovation and computing hardware advancement, the past decade has witnessed unprecedented success of deep neural networks (DNNs) across many real world applications such as image recognition, natural language processing, anomaly detection, driver-less cars, drones, etc~\cite{bojarski2016end,bourzac2016bringing,giusti2016machine,andor2016globally,graves2013speech}. However, 
Recent studies have shown that DNN models are inherently vulnerable to adversarial examples (AEs)~\cite{goodfellow2014explaining,szegedy2013intriguing}, i.e. malicious inputs crafted by adding small and human-imperceptible perturbations to normal inputs, strongly fooling the cognitive function of DNNs. 
%in speech recognition, it is possible to control user's devices like smart phone to visit a malicious webpage by playing a video with a crafted hidden voice command~\cite{}. 
For example, in image recognition, adversarially manipulating the perceptual systems of autonomous vehicles by physically captured adversarial images, i.e. via camera or sensor~\cite{ohn2016looking,smolyanskiy2017toward}, can lead to the misreading on road signs, thus causing potential disastrous consequences in DNN-based cyber-physical systems.

Many countermeasures~\cite{Liu:2018:SAE:3201607.3201772,liao2017defense,samangouei2018defense,song2017pixeldefend,tramer2017ensemble,akhtar2018threat} have been proposed to enhance the robustness of DNNs against adversarial examples, mainly including DNN model-specific hardening strategies and model-agnostic defenses~\cite{guo2017countering}.
%(refer to adversarial examples pre-processing techniques in this paper)~\cite{guo2017countering}. 
%However, these solutions either require expensive computation or show limited success against state-of-the-art attack benchmarks. 
Typical model-specific solutions like ``adversarial training" or ``defensive distillation" may rectify the model parameters to mitigate the attacks by using the iterative retraining procedure or masking adversarial gradient. 
%but they suffer from high training cost due to iterative retraining procedure. 
%``defensive distillation" is proved to be ineffective to counteract most recent Carlini \& Wagner attacks (or C\&W family attacks)~\cite{carlini2017towards}. 
The model-agnostic approaches such as input dimensionality reduction~\cite{bhagoji2017dimensionality,uesato2018adversarial} or direct JPEG compression~\cite{dziugaite2016study,das2017keeping,guo2017countering} attempt to remove adversarial perturbations from the inputs, before feeding them into DNN classifiers.
%appear too simple to sufficiently remove adversarial perturbations from input images without harming the DNN model accuracy~\cite{dziugaite2016study}. 
%\textcolor{red}{Introduce Quiling and TVM}~\cite{guo2017countering}
%However, recent study shows~\cite{} that the latest adaptive adversarial attack,

%However, most recent study~\cite{} shows that the latest adaptive adversarial attack, i.e., Backward Pass Differentiable Approximation (BPDA), can still successfully compromise these existing mitigation methods.

%, thus compromising the DNN-based machine learning.
%``Feature Squeezing", as one of the strongest model-agnostic defense techniques, is able to detect AEs with high accuracy and few false positives by accurately comparing the predictions of multiple models feeding with original samples and samples with feature squeezing ~\cite{xu2018feature}. 

%however, it is still very complicated because multiple models are needed in order to accurately compare the model prediction on the original sample with its predictions on the sample after feature squeezing~\cite{xu2018feature}.

\begin{figure*}[t]
 \centering
 \includegraphics[width=0.8\textwidth]{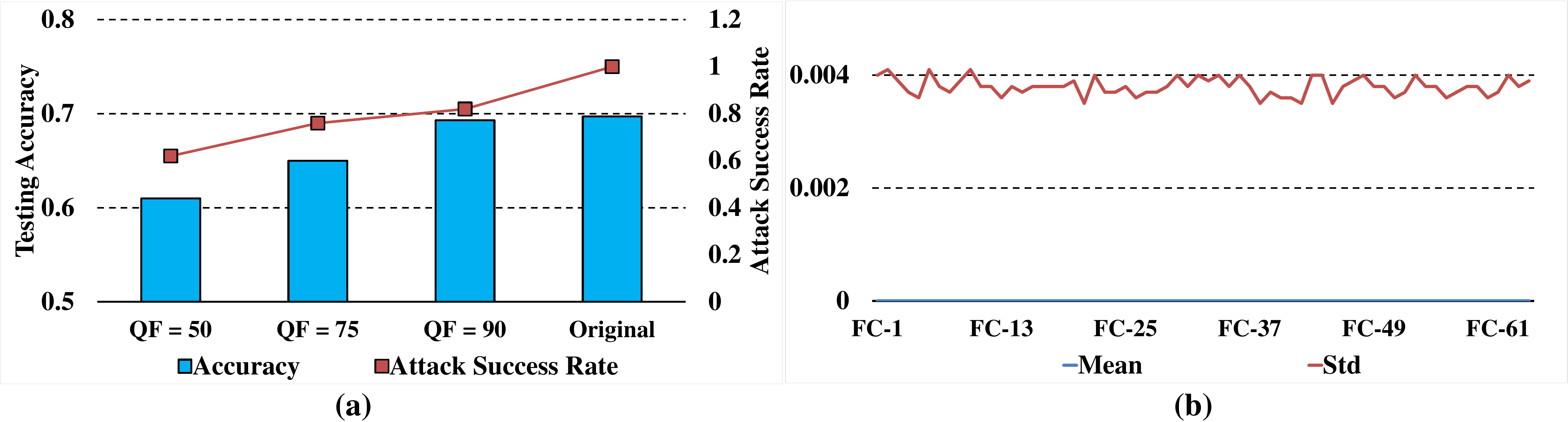}
 \caption{ \small (a) Testing accuracy v.s. attack success rate at different QFs of JPEG; (b) Statistical information of FGSM-based AE perturbations in frequency domain (FC denotes frequency component)}
 \vspace{-10pt}
 \label{acc}

\end{figure*}

In this work, we focus on improving the effectiveness and efficiency of compression based model-agnostic mitigation against adversarial examples.
%in image classification. 
Though standard JPEG compression has been explored to mitigate the adversarial examples~\cite{dziugaite2016study,das2017keeping}, it can neither effectively remove the adversarial perturbations, nor guarantee the classification accuracy on benign images, due to its focus on human visual quality.  
%its human-visual system (HVS) oriented design. 
%While directly deploying standard JPEG compression as a defense method~\cite{dziugaite2016study,das2017keeping} can neither effectively remove the adversarial perturbations nor guarantee the classification accuracy on benign images, 
Instead, we propose the DNN-favorable JPEG compression, namely \textit{``feature distillation"}, by redesigning the standard JPEG compression algorithm,
%redesign the JPEG compression framework, which is centered around human-visual system (HVS), to be both defensive and DNN-favorable,   
% (instead of centering around human-visual system (HVS)), 
% namely \textit{``feature distillation"},    
%augmented from standard JPEG, 
in order to maximize the defense efficiency while assuring the DNN testing accuracy. 
%Specifically
In specific, 1) We reveal the root reason to limit the JPEG defense efficiency by analyzing the frequency feature distributions of adversarial input perturbations during JPEG compression; 
2) Inspired by our observation, we propose a semi-analytical method to guide the defensive quantization process to maximize the effectiveness of filtering adversarial features; 
3) We characterize the importance of input features for DNNs by leveraging the statistical frequency component analysis within JPEG, and then develop DNN-oriented quantization method to recover the degraded accuracy (i.e., a side-effect induced by the feature loss in perturbation removal) on benign samples.

Our proposed method is built upon the light modifications of widely adopted JPEG compression and does not require any expensive model retraining or multiple model predictions.
Evaluations show that \textit{``feature distillation"} offers significantly improved effectiveness against a variety of mainstream adversarial examples (i.e., $>90\%$ accuracy on AEs), %including the strongest C\&W and latest BPDA attacks, 
with very marginal accuracy reduction (i.e., $\le1\%$) on benign data. Besides, it well beats recent proposed image transformation based defense like Quilting and TV Minimization in terms of defense efficiency, accuracy and processing speed. Furthermore, our solution offers the best defense efficiency ($\sim60\%$) with lowest accuracy loss ($\le1\%$) against the recent adaptive attack--Backward Pass Differentiable Approximation (BPDA)~\cite{athalye2018obfuscated} among existing input-transformation based defenses, though it is not completely immune to such attack.
%by $\sim4.5\times$ on effectiveness, 
%with $\sim260\times$ improvement on processing-time-per-image. 
%We also evaluate the defense efficiency of our method against recent adaptive adversarial attack--Backward Pass Differentiable Approximation (BPDA)~\cite{athalye2018obfuscated}.
%\textcolor{red}{I removed the following comment part}
% Compared with existing input-transformation techniques, our solution still offers the best defense efficiency ($\sim60\%$) with least accuracy reduction ($\le1\%$), although it is not completely immune to such attack.
\textit{To our best knowledge, there is no published work that can completely mitigate BPDA, since it is very challenging for defense if attackers can iteratively strengthen the adversarial examples according to the defense. However, we believe our work provides a new angle to redesign input-based defense to well balance the accuracy of benign data and defense efficiency with DNN-oriented/defensive quantization. It is a new trial towards developing better input-transformation based defenses.} 

%% file: 2.back.tex
\section{Background, Related Work and Motivation}
%JPEG~\cite{wallace1992jpeg} is a popular lossy compression standard for digital images. 
\subsection{Basics of Adversarial Examples and JPEG}
\textbf{Adversarial examples:} ($X^*=X+\delta_X$) are created to fool the DNNs ($Y^*\neq Y$) with imperceptible perturbations: %, which can be formulated as:
$\arg \min_{\delta_X}\parallel\delta_X\parallel\ s.t.\ F^{(\Theta)}\left( X+\delta_X \right) = Y^*$,
% \begin{equation}
% \arg \min_{\delta_X}\parallel\delta_X\parallel\ s.t.\ F^{(\Theta)}\left( X+\delta_X \right) = Y^*
% \end{equation}
which can be solved through many crafting algorithms:
%\begin{enumerate}
%\item
1) {\bf FGSM}~\cite{goodfellow2014explaining} (fast gradient sign method) %FGSM~\cite{goodfellow2014explaining} is optimized for $L_\infty$ distance matrix by crafting 
is a $L_\infty$ attack and utilizes the gradient of the loss function to determine the direction to modify all the input pixels. It is designed to be fast, rather than optimal;
%A small perturbation $\epsilon$ is added into all input pixels along the direction of the sign of the gradient: $X^{'} = X + \epsilon sign(\nabla_XL(X,Y))$,
%where $L$ is the loss function, $Y$ is the label and $\nabla_XL(X,Y)$ is the gradient of $L$ w.r.t. input $X$. Such a method is designed to be fast, rather than optimal.
%\item 
2) {\bf BIM}~\cite{kurakin2016adversarial} (basic iterative gradient sign method) is the iterative version of FGSM by gradually %\Nuo{gradually}
adding small perturbations $\alpha$ ($L_\infty$) 
%in iteratively 
until 
%augmented perturbation 
reaching the upper bound $\epsilon$ or achieving successful attack;
% :
% \begin{equation}
% \small
% X_0^{'}=X,X_{i}^{'} = X_{i-1}^{'} + clip_{\epsilon}\left( \alpha sign(\nabla_XL(X,Y))\right)
% \end{equation}
% Here the clipping equation, $clip_{\epsilon}(n)$, performs clipping on each pixel when it reaches $\epsilon$.
3) {\bf Deepfool}~\cite{moosavi2016deepfool} uses geometrical knowledge to search the minimal %adversarial 
perturbations ($L_2$)
% :
% \begin{equation}
% \small
% \triangle(X,X^{'}) := \arg\min_Z||Z||_2,~s.t.~F(X+Z)\neq F(X)
% \end{equation}
%In this method, 
by assuming DNN as
%DNN is treated as 
a linear classifier and each class is separated by a hyper-plane. Such an approach finds the nearest hyper-plane from $X$ and uses geometrical knowledge to calculate the projection distance;
4) {\bf C\&W}~\cite{carlini2017towards} (Carlini \& Wagner method) are a series of $L_0$, $L_2$, and $L_\infty$ attacks that achieve 100\% attack success rate with much lower distortions comparing with the above-mentioned attacks. In particular, the C\&W $L_2$ attack 
%is superior to L-BFGS attack because it 
uses a more effective objective function $f(x)=max\left(max\left\{Z(X)_i\ \ i\neq t\right\}-Z(X)_t,-\kappa \right)$ with logits $Z(X)_i$ and adjustable parameter $\kappa$. Further, C\&W $L_0$ and $L_\infty$ attacks are implemented indirectly by iteratively calling their $L_2$ attack.
%Backward Pass Differentiable Approximation (BPDA):athalye2018obfuscated
5) \textbf{BPDA}~\cite{athalye2018obfuscated} is the latest adaptive attack by recurrently computing the adversarial gradient after applying defense: 
$x^*=clip(x+\epsilon\cdot sgn(\triangledown_xJ_{\theta,Y}(DEF(x))))$, where $J$ represents the function of an DNN model and $DEF$ is the applied defense method in BPDA attack. It is state-of-the-art of attack by assuming adversaries know the defense method.

\textbf{JPEG:}~\cite{wallace1992jpeg} is a popular lossy compression standard for digital images based on the fact that Human-Visual System (HVS) is less sensitive to the high frequency components than low frequency ones~\cite{zhang2017just}. A typical JPEG compression mainly consists of image partitioning, discrete cosine transformation (DCT), quantization, zig-zag reordering and entropy coding, etc.~\cite{wallace1992jpeg}. 
To compress a raw image, the high (low) frequency DCT coefficients are usually scaled more (less) and then rounded to nearest integers by performing element-wise division based on a predefined $8\times8$ Quantization Table (Q-Table)~\cite{wallace1992jpeg}. The trade-off between image quality and compression rate is realized by scaling each element in Q-Table via the ``Quantization Factor" (QF)~\cite{ye2007detecting}. A higher compression rate corresponds to a lower QF. A reverse procedure of above steps can decompress an image.

% \textbf{The Carlini \& Wagner (CW) Method.}
% Carlini and Wagner~\cite{carlini2017towards} have proposed three methods to craft adversarial examples based on $L_0,L_2$ and $L_\infty$ norms as distance metrics. The adversarial agent searches for $w$ that is a new variable introduced by authors.  
% \begin{equation}
% \footnotesize
%  minimize \begin{bmatrix}\|\frac{1}{2}\left(tanh(w)+1\right)-X\|_p^2+c\cdot f\left(\frac{1}{2}(tanh(w)+1\right)\end{bmatrix}
% \end{equation}
% where $f(\cdot)$ is the objective function designed by authors based on the loss function:
% \begin{equation}
% \footnotesize
% f(x)=max\left(max\left\{Z(X)_i\ \ i\neq t\right\}-Z(X)_t,-\kappa \right)
% \end{equation}
% Here $Z(X)_i$ is the output of pre-softmax ``logit'' for class $i$. We can adjust $\kappa$ to control the confidence of adversarial attack success. Note that CW family attacks are the most recent and strongest attacks with least total distortion but can succeed in finding AEs for 100\% of images on defensively distilled networks~\cite{papernot2016distillation}. 

% {C\&W Attacks \cite{carlini2017towards}} are a series of $L_0$, $L_2$, and $L_\infty$ attacks that achieve 100\% attack success rate with much lower distortions comparing with the above-mentioned attacks. In particular, the C\&W $L_2$ attack is superior to L-BFGS attack because it uses a more effective objective function. However, C\&W $L_0$ and $L_\infty$ attacks are implemented indirectly by iteratively calling their $L_2$ attack.

\begin{figure*}[t]
 \centering
 \includegraphics[width=0.90\textwidth]{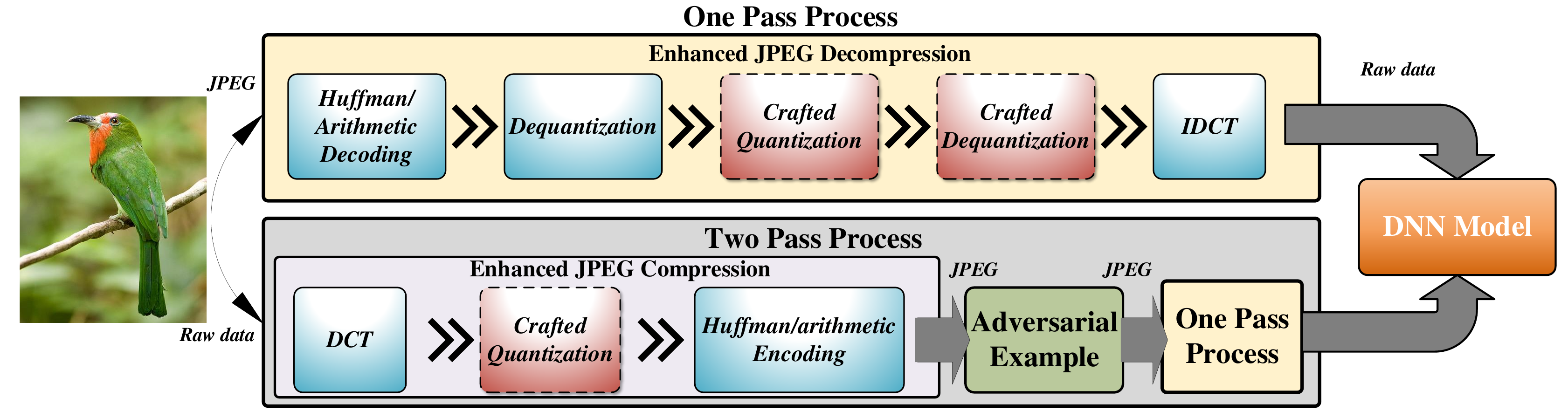}
 \caption{Illustration of two different modes of ``feature distillation"--one pass and two pass.}
 \label{fl}
 \vspace{-10pt}
\end{figure*}

\subsection{Related Works}
% \vspace{-5pt}
%\subsection{Related Works}
Applying JPEG compression to mitigate adversarial examples has been discussed in prior work.
%~\cite{kurakin2016adversarial,dziugaite2016study,guo2017countering,xu2018feature,das2017keeping,aydemir2018effects}.
Kurakin et al.~\cite{kurakin2016adversarial} 
%for the first 
test some model-agnostic approaches on adversarial examples
%, such as change of contrast and brightness, Gaussian blur, Gaussian noise, and JPEG compression, 
and reveal a good potential of JPEG compression for defending adversarial attacks. 
Dziugaite et al.~\cite{dziugaite2016study} empirically report JPEG compression can reverse only small adversarial perturbations, but the reason behind is uncertain. 
Guo et al.~\cite{guo2017countering} test JPEG compression, image Quilting (piecing together small patches from a database of image patches), total variance minimization (combining pixel dropout with total variation
minimization), etc. against
the gray-box and black-box
adversarial attacks,
%with gray-box and black-box 
%configurations, 
and report Quilting and TVM show better efficiency than JPEG.  
%of JPEG compression against gray-box attack.
Aydemir et al.~\cite{aydemir2018effects} compare the effects of JPEG compression and %its wavelet-based variant, 
JPEG2000, against adversarial perturbations.
Though JPEG2000 shows better performance than JPEG, the efficiency is still far from satisfactory.
%against adversarial attacks. 
Xu et al.~\cite{xu2018feature} propose an ensemble method, namely ``feature squeezing", to defeat the adversarial examples by integrating different types of ``squeezers'' (i.e. model-agnostic processing).
Das et al.~\cite{das2017keeping} propose a JPEG compression based ensemble method, namely ``vaccinating", to mitigate adversarial attacks by voting the result based on a variety of compression rates.
%\textbf{In summary, the prior JPEG compression xxxxxx; On the other side, our work xxxxx}
Prakash et al.~\cite{prakash2018deflecting} develop ``pixel deflection" and ``adaptive soft-thresholding" approaches by replacing or smoothing adversary perturbations. This method shows good defense efficiency on gray box-setting without evaluating adaptive attacks. 
%is not evaluated. 
Xie. et al.~\cite{xie2017mitigating} propose two randomization operations--random size and random padding, against adversary examples.
\textit{In summary, prior studies empirically test the JPEG compression by directly tuning the compression rate, 
%with different compression rate, 
without digging into
%analytics on 
the underlying image processing mechanisms.
%quantization mechanism. 
The conclusion is that JPEG suffers from very limited defense efficiency but inevitable DNN accuracy degradation. To overcome those issues, 
%Due to the limited efficiency 
%against adversarial attacks 
%and induced accuracy degradation, 
standard JPEG compression should be integrated with the costly ensemble solutions. On the other side, our work directly targets the fundamental entities of JPEG compression/decompression, like defensive and DNN-oriented quantizations, to unleash its defense potentials with almost zero loss of DNN testing accuracy, thus is low-cost.}

\vspace{-4pt}
\subsection{Why standard JPEG is not good?}
\label{mot}
%We find that, 
DNN suffers from both low testing accuracy and weak defense efficiency against adversarial examples if we directly employ standard JPEG compression based on human-visual system (HVS). To explore how existing compression can impact DNN's testing accuracy, we trained a MobileNet~\cite{howard2017mobilenets} with high quality JPEG images (QF=100, ImageNet), and tested it with both clear images and FGSM-based adversarial examples at various QFs (i.e., QF=100, 90, 75, 50). As Fig.~\ref{acc} (a) shows, the testing accuracy degrades significantly as the compression rate increases (or QF from 100 to 50), despite the slightly improved defense efficiency (or drop in attack success rate). To achieve the best defense efficiency among our selected four QFs (attack success rate = 0.62 at QF = 50), the accuracy is even reduced by $~\sim8\%$ on benign images than that of the original one (QF=100). Apparently, the HVS-based JPEG compression is not an ideal solution in terms of defense efficiency and accuracy.
%On the other hand, JPEG can only maintain the testing accuracy of DNN model at quite high QF values, but such less compressed JPEG cannot defend adversarial attacks effectively. 
%The next question is why default JPEG shows weak defense efficiency for AE attacks. Again We take the FGSM based AE to explore the perturbation distribution in the frequency domain. By following the JPEG compression procedure, we transfer the malicious perturbations to $8\times 8=64$ frequency components and analyze the corresponding statistical information. 
Fig.~\ref{acc} (b) further shows the means and standard deviations of DCT coefficients of malicious distortions at all 64 frequency bands. Given that malicious perturbations are almost randomly distributed in
every frequency band, HVS-based JPEG compression, which distorts more (less) on high (low) frequency components of the input, is unlikely to effectively filter the distortions across the whole spectral domain.

%Fig.~\ref{acc} (b) further shows the analysis in frequency domain, the means and standard deviations of the DCT coefficients of malicious perturbation at all 64 frequency components are almost the same. Therefore, the HVS-based JPEG compression, i.e., compressing more (less) on high (low) frequency components of an input image, is neither suitable for defending AE attacks nor for achieving high compression rate without accuracy loss. 

%% file: 3.design.tex
\section{Our Approach--Feature Distillation}
%Leveraging the JPEG compression technology as a defense method against AE attacks has been studied in previous works ~\cite{dziugaite2016study,das2017keeping,guo2017countering}, however, none of them have explained in details the defense principles and how to further optimize. 
In this section, we first provide a detailed analysis on how to 
%utilize the 
wisely redesign the quantization process in
JPEG compression to minimize attack success rate. As this lossy compression will still reduce the classification accuracy (see Fig.~\ref{acc}), we then develop the DNN-oriented quantization refine method, to compensate the reduced accuracy of benign images. Based on how/where the derived quantization will be placed in JPEG, our framework supports two modes (see Fig.~\ref{fl}): 1) \textbf{One pass process} by inserting a new quantization/de-quantization only in the decompression of standard JPEG; 2) \textbf{Two pass process} by also replacing the quantization of compression, followed by one pass process. The two pass method provides an opportunity to directly embed crafted quantization at sensor side to compress raw data to further improve defense efficiency, given that JPEG-based image compression, an integrated component in sensors, is usually a ``must-have" step to save data storage/transfer cost in real applications. \textit{Therefore, the one-pass handles incoming images compressed by standard JPEG before sending them to DNNs, while the two-pass targets raw data directly sampled by devices like image sensors.} 
%Since JPEG compression is usually an indispensable component within the sensing devices, the enhanced JPEG compression (with crafted quantization) in two pass process, can be naturally integrated at   
The target is to address both attack efficiency and test accuracy simultaneously.  
%both low attack efficiency and accuracy reduction issues. 
%Based on the number of

%\subsection{Analysis of JPEG Compression to Mitigate AE Attacks}
%\vspace{-7pt}
%\subsection{Analysis of Compression for Mitigating AEs}

%\textbf{Step 1: Defensive Quantization for Enhancing Defense Efficiency.}
\subsection{Step 1: Defensive Quantization for Enhancing Defense Efficiency}
\label{ae}
We propose to use spectral filter by leveraging quantization process in JPEG on DNN inputs (i.e., adversarial examples), in order to mitigate adversarial perturbations. 
%\st{The inputs with adversarial perturbations will be placed into ``DCT transformation'' and ``quantization'' processes in JPEG compression. }

\begin{figure*}[t]
\centering
\includegraphics[width=0.9\textwidth]{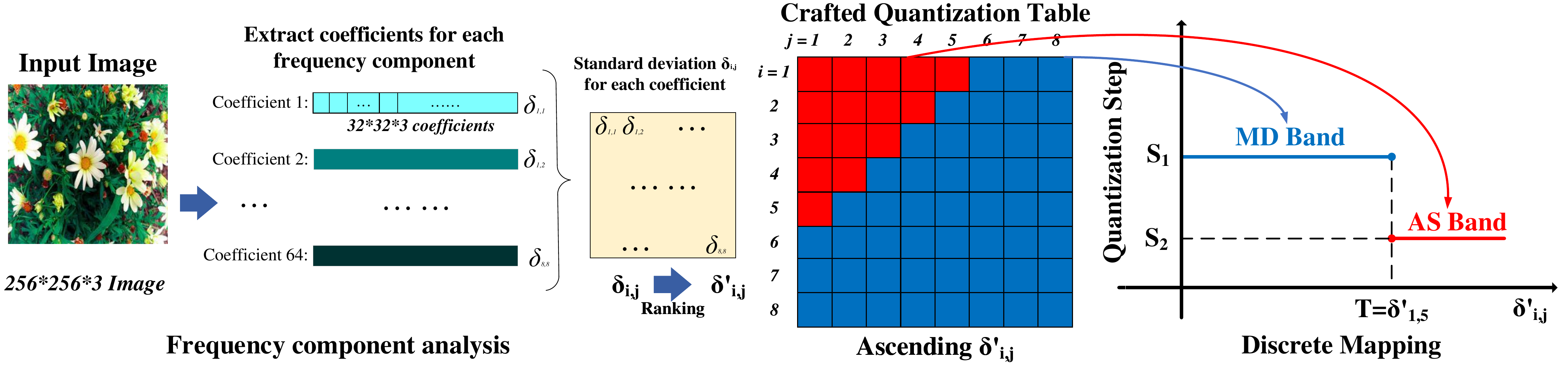}
\vspace{-5pt}
\caption{An overview of heuristic design flow of DNN-Oriented compression based on crafted quantization.}
\label{flowchart}
\vspace{-10pt}
\end{figure*}

%$X$ and $\delta X$
\textbf{One pass process.} The
%straightforward thinking
idea is to directly filter out the malicious perturbations in frequency domain through the quantization process. As Fig.~\ref{fl} shows, 
%before a agnostic JPEG form image feed into the DNN model
the JPEG-formatted input will be decompressed and then feed into the DNNs as the raw data at the beginning.
%it will first decompress the image into raw data. At this process 
By taking this chance, we 
%replace 
insert a new pair of quantization/dequantization processes after the dequantization of standard JPEG decompression 
%add a crafted quantization and dequantization process 
to purify the potential adversarial perturbations. Note we omit the first dequantization in the following analysis ideally by assuming it can almost preserve all frequency features of the input.
%preserves the image quality as much as possible. 
%We propose to redesign the quantization process of JPEG compression to directly filter the perturbations in frequency domain, with basic idea as follows: 
%For the second quantization, 
Assuming for each $8\times8$ block in the input image $X$, adversarial distortion $\delta_X$ is added to $X$ with intensity $\epsilon$. The DCT transformation--a linear operation, essentially projects the image from spatial domain to spectral domain. Therefore, the original input and adversarial perturbations could be linearly separated as: 
%Assuming for each $8\times8$ block in the input image $I$, perturbation $E$ is added to craft the adversarial example $I+E$, which can be generated through AE crafting algorithms with perturbation intensity $\epsilon$. 
%In JPEG compression, the step of DCT transformation will project the image from spatial domain to spectral domain by Discrete Cosine Transform (DCT), which is essentially a linear operation. As a result, the original input and adversarial perturbations could be linearly separated as: 
\begin{equation}
\footnotesize
DCT(X+\delta_X) = DCT(X)+DCT(\delta_X)=C_X\cdot B+C_{\delta_X}\cdot B
\end{equation}
where $C_X$ and $C_{\delta_X}$ are the DCT coefficients of $X$ and $\delta_X$, respectively, for the $8\times8$ image block, and $B$ is the DCT transformation basis.  The maximum magnitude of $C_{\delta_X}$ can be calculated by the summation of all 64 frequency components and each term is bounded by $cos(\theta)\cdot\epsilon$. Thus we have $-8\cdot\epsilon<C_{\delta_X}<8\cdot\epsilon$.
%Typically, following the DCT transformation, %\textcolor{red}{
%the maximum magnitude of $C_E$ can be calculated by the summation of all the 64 frequency components and each term is bounded by $cos(\theta)\cdot\epsilon$.  
%in spectral domain is defined as the accumulation of all the 64 frequency components bounded by $cos(\theta)\cdot\epsilon$
%}. 
%Thus we have $-8\cdot\epsilon<C_E<8\cdot\epsilon$.
% \begin{equation}
% \footnotesize
% -8\cdot\epsilon<C_E<8\cdot\epsilon
% \end{equation}
The DCT coefficients will be quantized again in this decompression process, providing a good opportunity for filtering the perturbations. The quantization is approximated as: 
\begin{equation}
\footnotesize
Round\left(\nicefrac{C_X + C_{\delta_X}}{QS}\right) \approx Round\left(\nicefrac{C_X}{QS}\right) + Round\left(\nicefrac{C_{\delta_X}}{QS}\right)
\label{Eq_CIQ}
\end{equation}
where $QS$ is the defensive quantization step (QS). Ideally, if $QS >\left | C_{\delta_X}\right |$, then the perturbation $C_{\delta_X}$ can be eliminated.
However, this equation may induce undesired rounding error to limit the efficiency of removing malicious perturbations, given that $C_{\delta_X}$ is usually much smaller than $C_X$.
%thus restrict the defense efficiency. 
We further analyze such a rounding error 
%let's disassembly  
%Through
by decomposing $C_X = \eta + \nicefrac{QS}{2}$, then we have: 
\begin{equation}
%\footnotesize
Round\left(\nicefrac{C_X + C_{\delta_X}}{QS}\right) = Round\left(\nicefrac{\eta + \nicefrac{QS}{2} + C_{\delta_X}}{QS}\right)
\label{Eq_CIQ1}
\end{equation}
If $\nicefrac{QS}{2} + C_{\delta_X} > QS$, this part will be rounded to $\pm1, \pm2,\pm3...$, which will 
induce a stronger rounding error than the adversarial perturbations, 
resulting in degraded defense efficiency.
%more benign features will be discarded along with the adversarial perturbations. 

\textbf{Two pass process.} To avoid
%this 
such rounding error, we further propose two pass method.
%decide 
%to quantize the legal inputs first 
%to filter the potential rounding error $\nicefrac{Q}{2}$ in benign inputs, 
As Fig.~\ref{fl} shows, %before the aforementioned one pass process,
%our design will first
the raw data (i.e. sampled by sensors) will be compressed through a defensive quantization process, rather than the standard JPEG quantization, followed by an entire one pass process.  
%we use the 1st pass (with the same proposed quantization design) 
%will be firstly applied 
%to generated the benign inputs by filtering the potential rounding error $\nicefrac{Q}{2}$ at the beginning, instead of the standard JPEG compression.
%distribute the 
%to filter the potential rounding error $\nicefrac{Q}{2}$ in benign inputs,
Assuming such compressed benign inputs are then polluted by adversarial perturbations, adversarial examples
%forward to DNN model, we quantize them again to purify the malicious perturbations
will be further processed by considering both compression/decompression procedures as:  
%through the 2nd pass (i.e., following the one pass method).
%as we mentioned before. 
%This
%Such a two pass process can be formulated as: 
\begin{equation}
Round\left(\nicefrac{(Round\left(\frac{\eta + \nicefrac{QS}{2}}{QS} \right) * QS + C_{\delta_X})}{QS} \right) = Round(\eta)
\label{Eq_CIQ2}
\end{equation}
%To this end, 
The malicious perturbations can be appropriately filtered without inducing any rounding error 
%To completely eliminate the perturbation
if $QS$ satisfies the following equation:
%$Round\left(\frac{C_E}{Q}\right) = 0$ 
%should be selected: 
\begin{equation}
\footnotesize
Round\left(\nicefrac{C_{\delta_X}}{QS}\right) = 0\Rightarrow QS > 2\left|C_{\delta_X}\right|,~C_{\delta_X}\in \left(-8\epsilon<C_{\delta_X}<8\epsilon\right)
\label{qval}
\end{equation}
%To this end, as long as the QS size 
Therefore, we adopt the same QS ($QS>16\cdot\epsilon$) to eliminate the perturbations $C_{\delta_X}$ in both passes.
\subsection{Step 2: DNN-Oriented Quantization for Compensating Accuracy Reduction}
\label{Analysis}
To recover the testing accuracy (see Section~\ref{mot}), our next step is to develop a DNN-oriented JPEG compression method by refining the defensive quantization table from step 1. We analyze the difference between human visual system (HVS) and 
DNN on feature extractions, and then propose a heuristic design flow. 
%to refine the quantization table.
%for mitigating accuracy loss. 

\textbf{Difference between HVS\&DNN on Feature Extractions.}
Since the feature loss happens in the frequency domain after the DCT process, we first study the problem that which frequency components have the most significant impact on DNN results.
Assume $x_k$ is a single pixel of a raw image $X$, and $x_k$ can be represented by $8\times8$ DCT: 
%in JPEG compression: 
\begin{equation}
%\footnotesize
x_k=\sum_{i=0} ^{7}\sum_ {j=0} ^{7} c_{(k,i,j)}\cdot b_{(i,j)}
\end{equation}
where $c_{(k,i,j)}$ and $b_{(i,j)}$ are the DCT coefficient and its basis function at 64 different frequencies, respectively.
It is well known that the human visual system (HVS) is less sensitive to high frequency components but more sensitive to low frequency ones. The JPEG quantization table is designed based on this fundamental understanding. However, DNNs examine the importance of the frequency information in a quite different way. The gradient of the DNN function $F$ with respect to a basis function $b_{(i,j)}$ is calculated as:
\begin{equation}
%\footnotesize
{ \nicefrac{\partial F}{\partial b_{(i,j)}}=\nicefrac{\partial F}{\partial x_k}\times\nicefrac{\partial x_k}{\partial b_{i,j}}=\nicefrac{\partial F}{\partial x_k}\times c_{(k,i,j)} }
\label{basis}
\end{equation}
% \begin{equation}
% %\footnotesize
% { \frac{\partial F}{\partial b_{(i,j)}}=\frac{\partial F}{\partial x_k}\times\frac{\partial x_k}{\partial b_{i,j}}=\frac{\partial F}{\partial x_k}\times c_{(k,i,j)} }
% \label{basis}
% \end{equation}

Eq.~(\ref{basis}) implies that the contribution of a frequency component ($b_{i,j}$) to the DNN result will be mainly decided by its associated DCT coefficient ($c_{(k,i,j)}$) and the importance of the pixel ($\nicefrac{\partial F}{\partial x_k}$). Here 
%$\nicefrac{\partial F}{\partial x_k}$ is obtained after the DNN training, while 
$c_{(k,i,j)}$ will be distorted by the quantization before training. Ideally a well trained DNN model should respond with different strengths to all the 64 frequency components depending on the $c_{(k,i,j)}$ values. From this observation, large $c_{(k,i,j)}$ should be compressed less (using a small quantization step) in order to ensure a desirable classification accuracy. 

In contrast, the default quantization table used in JPEG focuses on compressing more on less sensitive frequency components to HVS. As a result, in order to defend against adversarial attacks, aggressive compression is required, making DNNs easily misclassified if original versions contain important high frequency features.
%In CASE 1 (see Fig.~\ref{acc}(a)), 
The DNN models trained with original images learn comprehensive features, especially high frequency ones. However, such features are actually lost in more compressed testing images, resulting in considerable misclassification rate (see Fig.~\ref{acc}(a)). 
%Therefore, directly using the JPEG compression method as AE defense is not an optimal solution.

\iffalse
\begin{figure*}[t]
\centering
\includegraphics[width=0.8\textwidth]{figures/qsize}
%\vspace{-10pt}
\caption{The impact of quantization table size on various AE attacks at different quantization steps.}
\vspace{-10pt}
\label{qsize}
\end{figure*}
\fi

Therefore, we propose to compensate the accuracy reduction incurred by defending adversarial examples through a heuristic design flow (see Fig.~\ref{flowchart}): 
1) characterize the importance of each frequency component through frequency analysis on benign images; 2) lower the quantization step of the most sensitive frequency components based on the statistical information for enhancing accuracy. 

% \begin{algorithm}[t]
% \small
% \caption{\label{alg_2}Frequency Component Analysis Algorithm}
% \DontPrintSemicolon
% %\tcp{Pseudocode of Frequency component analysis Algorithm}%: $V_{max}$/$T_{max}$ is recorded.}
% $fimg$: Image in frequency domain;\;
% $fc_k$: Frequency components;\;
% Nblock: \# of 8*8 blocks after block-wise DCT;\;
% $\delta_k$: standard deviation of kth frequency components;\;

% $fimg$ = 8*8 block-wise DCT ($img$)\;
%    \ForEach{$Block_{i,j}$ in [1 .. Nblock]} {
%     $Block_{i,j}$ = $fimg_i$[j*8-8:j*8][j*8-8:j*8]\tcp{i-th sampled image j-th 8*8 block}
%       \ForEach{$fc_k$ in [1 .. 64]}{
%          $fc_k$ store $Block_{i,j}[k]$\tcp{i-th sampled image j-th 8*8 block k-th frequency component}
%     }
%   }  

% \tcp{Statistical Analysis}
% \ForEach{$fc_k$ in [1 .. 64]}{
% calculate standard deviation $\delta_k$}
% return $\delta_k$ \tcp{standard deviation of each frequency component}
% \end{algorithm}

% \begin{figure}[b]
% \centering
% \includegraphics[width=0.9\columnwidth]{figures/topn}
% \vspace{-10pt}
% \caption{Impact of quantization step on $acc_{l}$ and $acc_{m}$ .}
% \label{topn}
% \end{figure} 

%\vspace{-5pt}
\textbf{A: Frequency Component Analysis.}
For each input image, we first characterize the pre-quantized DCT coefficient distribution at each frequency component. Such a distribution represents the energy contribution of each frequency band~\cite{reininger1983distributions}. Prior work~\cite{reininger1983distributions} has proved that the pre-quantized coefficients can be approximated as normal (or Laplace) distribution with zero mean but different standard deviations ($\delta_{i,j}$). A larger $\delta_{i,j}$ means more energy in band $(i,j)$, hence more important features for DNN learning. As Fig.~\ref{flowchart} shows, each image will be first partitioned into $N$ $8\times8$ blocks, followed by a block-wise DCT. Then the DCT coefficient distribution at each frequency component will be characterized by sorting all coefficients at the same frequency component across all image blocks. The statistical information, such as the standard deviation $\delta_{i,j}$ of each coefficient, will be calibrated from each individual histogram.

\textbf{B: Quantization Table Refinement.}
Once the importance of frequency components is identified based on the standard deviations of DCT coefficients ($\delta_{i,j}$), the next step is to boost the accuracy of legitimate examples $\left\{acc_{l}\right\}$ (refer to the testing accuracy of benign images processed after the defense method).  
%to boost the accuracy. 
%The basic idea is to utilize finer quantizations at the important components by leveraging the intrinsic error resilience property of DNNs. 
Our analysis in Section~\ref{mot} indicates that a proper selection of $QS$ can effectively mitigate the perturbations, whereas larger $QS$ will induce more quantization errors. Therefore, we reduce the quantization errors of the most sensitive frequency components to enhance the testing accuracy by lowering their corresponding quantization steps within the quantization table, but such frequency components should be as few as possible to maintain the defense efficiency.
% Note the table composed of the same QS values ($QS=30$) delivers the best AE defense efficiency. 
%To maximize the testing accuracy without impacting the defense efficiency, 
%we focus on minimizing the quantization errors on those most important frequency components (but as few as possible).
%will be reduced. 
%Note we must apply quantize process on the most important frequency components, since these components also contain malicious perturbations which will significantly impact the defense efficiency.  
% \begin{equation}
% \label{opt}
% max\left\{acc_{l}\right\}\, s.t.\left\{\# fc_s\right\}
% \end{equation}
% where $\# fc_s$ is the number of most sensitive frequency components. 
In specific, we first sort the magnitude of $\delta_{i,j}$ in an ascending order as $\delta^{'}_{i,j}$, then set the appropriate quantization step based on $\delta^{'}_{i,j}$.
%for both   $QS=20$ and $QS=30$ for $top-n$ largest $\delta^{'}_{i,j}$ and the remained $\delta^{'}_{i,j}s$, respectively. 
%From our experimental results $QS_2=20$ at the position of the top 15 frequency components can ensure the testing accuracy degradation less than $\sim1\%$.
% The relationship between $n$ and $acc_l$ shown in Fig.~\ref{topn} indicates that set the .
%As Fig.~\ref{topn} shows, 
%to 1 
%which means no quantization at those frequency components, 
%meanwhile keep the other $QS$ all 40. 
% Consequently, we divide the different frequency components into two bands: \textit{accuracy sensitive (AS) band} and \textit{malicious defense (MD) band}. %For convenience
To simplify our design, we introduce a discrete mapping function
%(DM) 
to derive the quantization step on each frequency band, base on the associated standard deviation $\delta_{i,j}$, i.e.,
%(see Fig.~\ref{flowchart})
$QS_{i,j}=(\delta_{i,j}\leq T\text{ ? }S_1\text{ : }S_2)$, 
% \begin{equation}
% \footnotesize
% \label{plm}
% Q_{i,j}=\begin{cases}S_1 & \delta_{i,j} \leq T\\S_2 & \delta_{i,j}>T \end{cases}
% \end{equation}
where
% $QS_{i,j}$ is the quantization step at the frequency band $(i,j)$ and 
$T$ is the threshold to divide the 64 frequency components.
%bands according to ascending order of the magnitude of $\delta_{i,j}$. 
Note that $S_1>S_2$.
%As right part of Fig.~\ref{flowchart} shows, 
The 64 frequency components are divided into two bands (see Fig.~\ref{flowchart}): the red colored Accuracy Sensitive (AS) band with $QS=S_2$,
%, consists of 15 largest $\delta^{'}_{i,j}$
and the blue colored Malicious Defense (MD) band with $QS=S_1$ from Section~\ref{ae}.
%, consists of the others. 

%Hence, we adopt $T = \delta^{'}_{1,5}$, $S_1=30$, and $S_2=20$ in our design. 
%Two different QS  $S_1$ and $S_2$ are defend as 1 and 40 from our experimental results. 

%% file: 4.evaluation.tex
\begin{figure*}[t]
\centering
%\vspace{-20pt}
\includegraphics[width=0.9\textwidth]{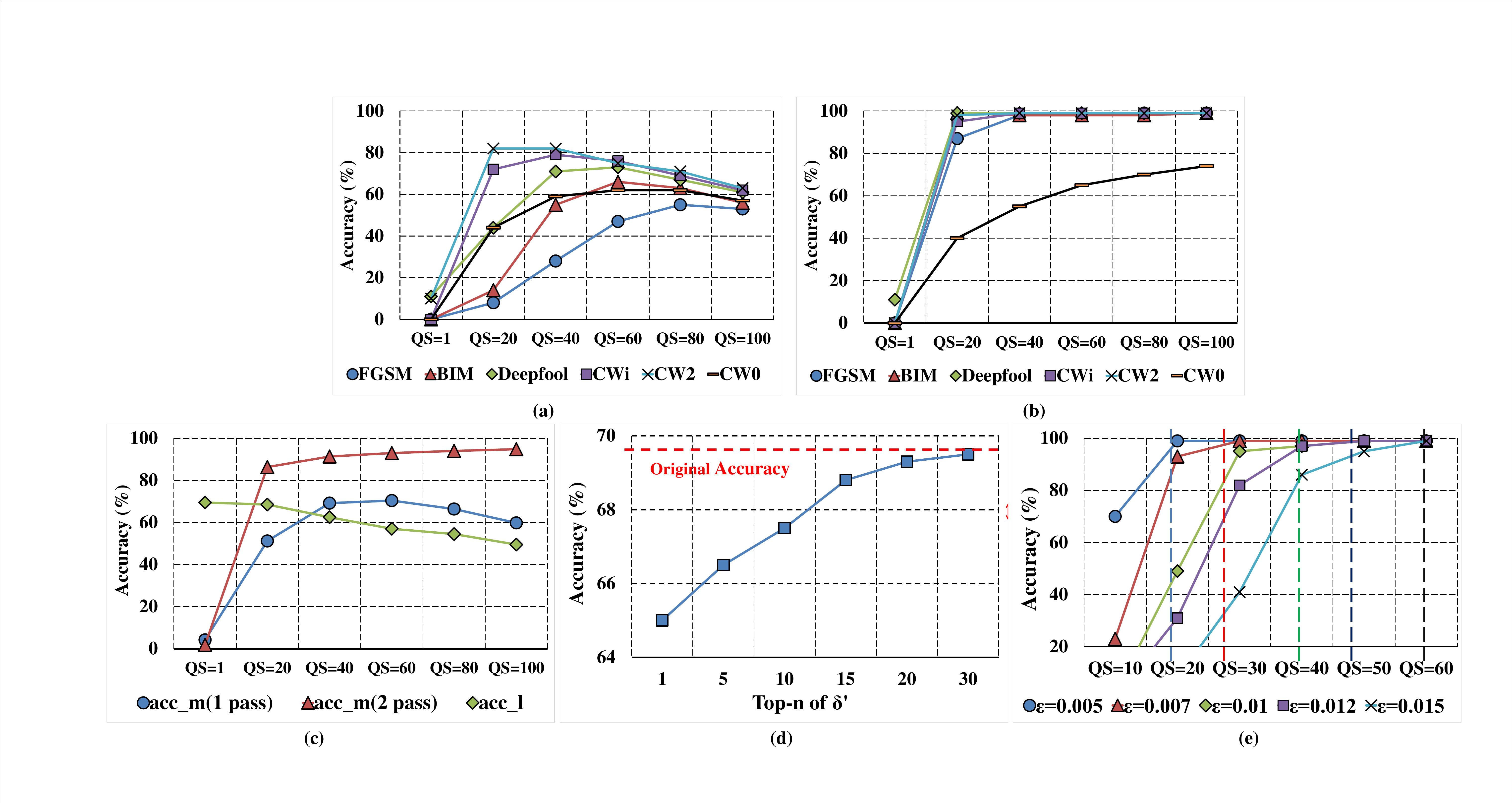}
\vspace{-14pt}
\caption{Exploration of the defensive quantization step for a 8*8 table: 
(a) Defense efficiency of one pass method against adversarial examples; 
(b) Defense efficiency of two pass method against adversarial examples; 
(c) Average defense efficiency w.r.t. the legitimate image accuracy (FGSM, $\epsilon=0.008$); 
(d) Accuracy impacts of ranked frequency components (FGSM, $\epsilon=0.008$); 
(e) Accuracy impacts of various quantization steps w.r.t. different perturbation strength (FGSM).}
\vspace{-10pt}
\label{qs}
\end{figure*}

\section{Evaluation}
In this section, we first explore the parameter optimization in our feature distillation under the constraints of high classification accuracy on malicious inputs after applying defense, 
%In this section we first optimize the parameters in our ``feature distillation" with the constraints of high classification accuracy on malicious inputs after defense applied,
%, since any practical defense should well handle malicious samples but 
while preserving the accuracy of legitimate ones given that both types of data can arrive for a realistic DNN testing.
%We evaluate the robustness against adversarial perturbations of ``feature distillation" with the constraint of high classification accuracy on legitimate inputs, since any practical defense should well handle malicious samples but at the same time not impact the accuracy of legitimate ones given that both types of data can arrive for a realistic DNN testing. 
%\textbf{Threat Model}
Then we comprehensively evaluate feature distillation under following three different %configurations
%on threat models
%by extending assumptions from~\cite{guo2017countering}
%--gray box~\ref{black}, black box~\ref{gray} and white box~\ref{white}, with the following 
settings:
%~\cite{athalye2018obfuscated,das2018shield,guo2017countering}
1). \textbf{Gray-box:} We assume the adversary has full access to DNN model, but is unaware of the input transformations applied (defense method unaware)~\cite{guo2017countering,das2018shield}.	
2). \textbf{White-box:} We consider adversary has full access to the DNN model, as well as the full knowledge of the defense method~\cite{athalye2018obfuscated}, which is more challenging.
3). \textbf{Black-box:} We assume adversary does not know the exact network architecture and weights, instead, can use a substitute model to craft adversarial perturbations that are transferable to the target model~\cite{guo2017countering}.

%1) Section~\ref{black} Black Box setting; 2). Section~\ref{gray} Gray Box setting; 3). Section~\ref{white} White Box setting, for a comprehensive evaluation. 
%Then we implement our proposed ``feature distillation'' method in three different threat model types, 1). Section~\ref{black} Black Box setting; 2). Section~\ref{gray} Gray Box setting; 3). Section~\ref{white} White Box setting, for a comprehensive evaluation. 
%Our defense method more focus on gray box setting, however the adaptive attack (BPDA) and black box results are also evaluated. 

\subsection{Experimental Setup}
Our experiments are conducted on the Tensorflow DNN computing
framework~\cite{abadi2016tensorflow}, running with Intel(R) Xeon(R) 3.5GHz CPU and two parallel GeForce GTX 1080Ti GPUs. Our proposed feature distillation method is implemented on the heavily modified adversarial machine learning library--EvadeML-Zoo~\cite{xu2018feature} for white and gray-box settings and BPDA attack~\cite{athalye2018obfuscated} for white box setting.
%Obfuscated Gradients~\cite{obfuscated-gradients} for white box setting (BPDA).
To better illustrate the image compression based mitigation, we choose the large-scaled ImageNet dataset as our benchmark. Four other input-based countermeasures, including default JPEG~\cite{dziugaite2016study,kurakin2016adversarial}, bit-depth (one of the feature squeezing methods by reducing the bit number of an image pixel) ~\cite{xu2018feature} and the recent proposed TV Minimization (TVM) and Image Quilting~\cite{guo2017countering}
%countermeasures from~\cite{guo2017countering}--the 
, are selected as the baselines to compare with our proposed feature distillation. 
%-- a benchmarking and visualization tool for adversarial machine learning.
%Since our defense approach is based on image compressions technology,

%Two DNN models, ``MobileNet'' and ``VGG-19'', are selected to evaluate the defense efficiency.
%Two other defense methods, i.e. default JPEG~\cite{dziugaite2016study,kurakin2016adversarial} and ``feature squeezing''~\cite{xu2018feature}, are selected as the baselines to compare with proposed ``feature distillation". 

\textbf{Methodology.} 
%We assume the attacker has the full knowledge on target DNN models.
Various types of adversarial example attacks,
%candidates,
i.e., FGSM, BIM, Deepfool, CW$_2$, CW$_0$, CW$_\infty$ and adaptive attack--BPDA, 
have been simulated in our experiments for evaluating the defense.
%countermeasures.
%by following the same configuration and perturbation strength in ``feature squeezing"~\cite{xu2018feature}.
%By leveraging the evaluation models used by~\cite{xu2018feature},
We adopt a similar evaluation model from~\cite{xu2018feature}.
First, we choose 1000 benign images (one per class) to evaluate the testing accuracy of each DNN model.
The seed images, which will be adding adversarial  perturbations, are selected from the first 100 correctly predicted examples in the 1000 selected images on each DNN model for all the attack methods. 
The legitimate examples \textit{classification accuracy} ($acc_l$) is the testing accuracy of benign images processed by the defense method. 
The \textit{defense efficiency} is measured by the classification accuracy ($acc_m$) of 100 polluted images after applying the defense method.

\begin{table*}[t]
\centering
 \caption{The defense efficiency (classification accuracy on adversarial examples) of selected defense methods against different adversarial attacks.}
 %\caption{Mitigation efficiency (classification accuracy (\%) under adversarial attacks) of selected defense methods against different adversarial attacks.}
 \label{t1}
\vspace{-10pt}
\begin{tabularx}{\textwidth}{cccccccccc}
\Xhline{3\arrayrulewidth}
                  & \textbf{FGSM}        & \textbf{BIM}         & \textbf{DeepFool}    & \textbf{CW2}         & \textbf{CW0}         & \textbf{CWi}         & \textbf{Average}     & \textbf{acc$_l$}   & \textbf{Time (s)}     \\
\Xhline{2\arrayrulewidth}
\textbf{No defense (\%)}        & 0           & 0           & 11          & 10          & 0           & 0           & 3.5         & 69.5     & 0.11     \\
\textbf{Bit-depth (5-bit) (\%)}         & 2           & 0           & 21          & 68          & 7           & 33          & 21.83       & \textbf{69.4} & 0.04 \\
\textbf{JPG (90) (\%)}         & 5           & 9           & 9           & 68          & 5           & 32          & 21.33       & 69            & 0.11\\
\textbf{Quilting (\%)}          & 48          & 61          & 47          & 50          & 48          & 49          & 50.5        & 63.5   &  32.47     \\
\textbf{TVM (\%)}               & 33          & 42          & 68          & 77          & 49          & 90          & 59.8        & 60       &   38.89  \\
\Xhline{2\arrayrulewidth}
\textbf{FD-1P (\%)}             & 13           & 35          & 63          & 86          & 61          & 78          & 56      & 68.5    & 0.16     \\
\textbf{FD-2P (\%)}             & \textbf{92} & \textbf{99} & \textbf{99} & \textbf{99} & 58 & \textbf{99} & \textbf{91} & 68.5      & 0.16    \\
\Xhline{3\arrayrulewidth}
\end{tabularx}
\vspace{-12pt}
\end{table*}

\vspace{-3pt}
\subsection{Optimized Quantization Step}% Exploration}
\label{exp}
\vspace{-3pt}

%{\bf Defending against adversarial examples.}
{\bf Defending against adversarial examples.}
% \iffalse
% Based on our quantization design (in Section~\ref{Analysis}), 
% we first evaluate the defensive quantization step ($S_1$) by exploring the impact of different quantization table sizes (i.e., 4*4, 8*8, 32*32, 112*112, 224*224) against various adversarial attacks.
% Note here we assume $S_1=S_2$ to evaluate the best achievable defense efficiency.
% %We chose 5 different sizes (i.e. 4*4, 8*8, 32*32, 112*112, 224*224) 
% As Fig.~\ref{qsize} shows, for a smaller QS ($\leq 40$), different table sizes do not give too much differences on the classification accuracy of adversarial examples. However, as QS increases ($\geq 60$), small table size (i.e., 4*4 and 8*8) alway achieves higher accuracy than the larger ones, indicating better defense efficiency.
% Therefore, we use the same quantization table size (8*8 ) as that of standard JPEG, to secure the defense efficiency and reduce the computational cost.     
% \fi
%As we mentioned in previous section,
%first, large QS will significantly degrade the legitimate image testing accuracy; second,small QS is insufficient to against AEs. Therefore we keep use the default 8*8 table size.
Fig.~\ref{qs} (a) and (b) illustrate the impact of the quantization steps of the 8*8 table under various adversarial attacks with our one-pass and two-pass defense approaches applied, respectively. Apparently, both processes demonstrate that the defense efficiency can be steadily improved as the QS grows, however, it will be saturated (even decreased) if QS becomes too large for the two pass (one pass) process. 
%We can see first all the attacks have significantly improvement by leverage two pass method compare with one pass method, except CWl0 attack. 
Compared with the one pass process, the two pass process always delivers much better defense efficiency against most of the adversarial attacks (except the CW$_0$), 
due to the 
%The improvement is caused by the 
elimination of the rounding error. 
%However L0 attack don't have too much difference, this is because L0 attack is target to 
The reason is because CW$_0$ attacks attempt to use a minimum number of pixel(s) with maximum perturbations to fool the DNN models, therefore the perturbations of each single pixel will translate into larger magnitudes than the other attacks in the frequency domain. This leads to a much higher QS for completely removing the associated perturbations, as Fig.~\ref{qs} (a) and (b) show.
%in frequency domain after DCT. 
%Therefore, enlarging the QS (with reduced image quality) can relieve such an attack by filtering part of the perturbations.
%Therefore, only enlarge the QS can relieve this attack, as shown in Fig.~\ref{qs} (a) and (b) as QS increase the defense efficiency of L0 is improved. 

{\bf Evaluating testing accuracy.} Fig.~\ref{qs} (c) shows the testing accuracy changes w.r.t. QSs for both malicious examples ($acc_m$) and legitimate examples ($acc_l$).
%degrades with QS increasing, especially when QS larger than 20.
%Here 
The $acc_m$ (1 pass) and $acc_m$ (2 pass) represent the average accuracy of various adversarial examples by applying our one pass and two pass process, respectively. 
%with different QS values, 
%and acc\_l indicates the testing accuracy on legitimate images. 
%As shown in Fig.~\ref{qs} (c), $acc_{m} (1 pass)$ curve and $acc_l$ curve at first demonstrate an opposite trend and have a cross-over at $QS = 20-40$. 
As Fig.~\ref{qs} (c) shows, $acc_m$ (1 pass) and $acc_l$ demonstrate an opposite trend as QS grows, but they have a cross-over zone between QS=20 and QS=40. The adversarial perturbation dominates the accuracy reduction before the cross-over point (small QS), however, after that, both $acc_m$ (1 pass) and $acc_l$ will decrease due to the enlarged QS. On the other hand, $acc_m$ (2 pass) increases consistently as QS increases because of additional defensive quantization in compression stage.
{\it Therefore, we set $S_2$=20 and $S_1$=30 for the top-n largest $\delta^{'}_{i,j}$ (AS Band) and the others (MD Band), respectively, to  better balance the $acc_m$ and $acc_l$, according to our flow in Fig.~\ref{fl}.
Fig.~\ref{qs} (d) validates that such a configuration of $(S_1,S_2)$ at $n=15$ minimizes the degradation of $acc_l$ ($\le1\%$).} 

\textbf{Theoretical validation of defensive QS.} Fig.~\ref{qs} (e) further compare our analytic results (see Eq.~\ref{qval}) with experimental results for selecting QS.
%We select the FGSM attack to explore the theoretical results of our method and the experimental results in real case. 
We use FGSM attack with 5 different perturbation strengths (i.e. $\epsilon = 0.005, 0.007, 0.01, 0.012, 0.015$) as an example. The corresponding analytic QS values based on Eq.~\ref{qval} should be: 20.5, 28.7, 41, 49.2 and 61.44, respectively (dash lines in Fig.~\ref{qs} (e)). As expected, those analytical values are in excellent agreement with the experimental results when the defense efficiencies reach 100\%.  

%to achieve the best defense efficiency ($\sim100\%$ accuracy), the actual QS value in our experiments is similar to the analytic value for each given perturbation strength. These results verified our proposed method.
%every perturbation strengths achieve the 100\% defense efficiency (or accuracy) nearby the theoretical required QS values. This results further verify our method. 
% perturbation strength 

\begin{table*}[t]
\centering
 \caption{The defense efficiency (accuracy on adversarial examples--$acc_m$ and accuracy on legitimate images--$acc_l$ on ImageNet, against adaptive adversarial attack--BPDA.}
 \label{t2}
\begin{tabularx}{\textwidth}{ccccccccccc}
\Xhline{3\arrayrulewidth}
       & \textbf{None} & \textbf{Bit-depth} & \textbf{Quilting} & \textbf{TVM} & \textbf{JPEG(75)} & \textbf{JPEG(20)} & \textbf{JPEG(10)} & \textbf{FD(1x)} & \textbf{FD(2x)} & \textbf{FD(3x)} \\
\Xhline{3\arrayrulewidth}
\textbf{$\boldsymbol{acc_m}$ (\%)} & 0 & 0         & 0        & 0   & 0        & 34       & 45       & 10     &  42     & 60    \\
\textbf{$\boldsymbol{acc_l}$ (\%)} &  78  & 77        & 72       & 68  & 74       & 68       & 61       & 77     & 76     & 74     \\
\Xhline{3\arrayrulewidth}
\end{tabularx}
\vspace{-15pt}
\end{table*}

% \vspace{-8pt}
\subsection{Enhanced Robustness Against AE}
Based on our explorations on parameters optimization in section~\ref{exp}, we adopt $S_1=30, S_2=20, n=15$ to evaluate the overall defense efficiency. Note although we focus on defense efficiency, the images compressed by our method still provide acceptable visual quality (detailed results are summarized in the supplemental material). 

%\textcolor{red}{(Visual results under our compression method are summarized in the supplemental material).}
%We following previous works, evaluate our defense method in three threat models:

\vspace{-10pt}
\subsubsection{Gray Box Mitigation}
\label{gray}
%  \vspace{-8pt}
%\textcolor{red}{Measurement goes here. To evaluate the mitigation efficiency, we measure the classification accuracy on XXX with models XXX.}
Table.~\ref{t1} compares the defense efficiency of two proposed methods (i.e., 1-pass feature distillation \textbf{FD-1P} and 2-pass version \textbf{FD-2P})
%for ``feature distillation'' one pass design and ``FD-2P'' for two pass design) 
with five baselines--no defense, JPEG, Bit-depth, Quilting and TVM against 6 selected adversarial examples for MobileNet. 
%All the experiment setting is the same as previous work~\cite{xu2018feature}. 
\textit{Note that JPEG (90\% quality) and Bit-depth (5-bits) are conducted under the premise of $\le1\%$ legitimate classification accuracy reduction. However, the other two methods Quilting and TVM, cannot satisfy this constraint, so we compare our approach with those two methods on both $acc_l$ and $acc_m$.}  

%and ``VGG-19". 
%We first evaluate the defense efficiency 

\textbf{Comparison with bit-depth and JPEG.} We first limit our comparisons to the defense with $\le1\%$ reduction of $acc_l$ under no defense. In this case, Quilting and TVM are not included and will be compared separately. 
%with $\le1\%$ reduction of the $acc_l$ under no defense. 

Overall, FD-2P shows much better performance than that of FD-1P (56\% v.s. 91\% on average).
%for both models . 
%evaluates the enhanced robustness on target DNN models by comparing the defense efficiencies of our proposed method (i.e., ``FD-1P'' for ``feature distillation'' one pass design and ``FD-2P'' for two pass design) with three other baselines (include ``no defense'') across various adversarial attacks on ImageNet dataset. 
%It can be observed that:
Compared with no defense baseline,
%without applying any defense mechanisms, 
our FD-2P improves the average accuracy on adversarial examples from $\sim3.5\%$ to $\sim91\%$, which demonstrates the best mitigation efficiency among all methods.
%both ``MobileNet'' and ``VGG-19''. Moreover, 
% The defense techniques satisfy the aforesaid accuracy requirement on legitimate samples. The attack types, distortions, and success rates on the original images (without defense) are the same as Table 2. 
% As shown in the figures, 
Moreover, FD-2P can significantly outperform two other defensive baselines among all selected adversarial examples, i.e. improved by $\sim69\%$ (or $\sim73\%$) on average than the bit-dept (5-bit) or JPEG on both DNN models.

%CW$_\infty$
Particularly, for $L_\infty$ attacks like FGSM, BIM and CWi, existing model-agnostic methods show very limited efficiency. Similarly, our one pass method FD-1P shows marginal improvement when compared with the existing approaches. However, our two pass method FD-2P can almost completely remove this type of $L_\infty$ perturbations 
%in such kind of adversarial examples, 
and deliver the best defense efficiency. 
Besides, for the $L_2$ attacks, especially CW$_2$, existing defense methods show good defense efficiency ($\sim68\%$). Again FD-2P can rectify this kind of adversarial examples with almost 100\%. 
Compared with $L_\infty$ and $L_2$, the improvement of $L_0$ attacks ($CW_0$) is less significant, however, 
%performs the most defense efficiency with the reason we have discussed before. Particularly,our method achieve much improvement compare with the other two methods. 
%To better defense this type of attacks, we can implement other methods, like remove the odd values in a small spatial window.
%our method achieves $\sim70\%$ more defense efficiency improvement than all two other defense methods on average. 
FD-2P still achieves more than $50\%$ defense efficiency improvement comparing with bit-depth and JPEG. That is because, JPEG (90\% quality) uses small quantization steps (or large QFs) to maintain the quality of legitimate images for desirable accuracy, however, is also resulting in a low defense efficiency. Bit-depth roughly quantizes all image pixels uniformly, while our method distills the features in a more fine-grained manner by maximizing the loss of adversarial perturbations and minimizing the distortions of benign features. 

\textbf{Comparison with Quilting/TVM.} We also compare our solutions with Quilting and TVM in three aspects: $acc_m$, $acc_l$, and processing-time-per-image. Our average defense efficiency is much higher than the other two, i.e. 56\%/91\% (FD-1P/FD-2P) v.s. 50.5\% (quilting), 59.8\% (TVM). 
We also achieve the best testing accuracy ($acc_l$), that is 68.5\% (FD-1P/2P) v.s. 63.5\% (quilting), 60\% (TVM). 
%These legitimate accuracy reduction, i.e. 6\% (Quilting) and 9.5\% (TVM) is unacceptable.
Moreover, 
%these two methods are time costly, for the processing-time-per-image, 
we improve the processing-time-per-image (i.e., 0.15s on FD-1P) by $\sim216\times$ ($\sim259\times$) compared with Quilting (32.4s) and TVM (38.8s), or 0.15s (FD-1P) v.s. 32.4s (quilting) and 38.8s (TVM), as Table~\ref{t1} shows.

%Moreover, 
In general, our proposed feature distillation is particularly effective to mitigate stronger attacks (i.e. CW attacks with least perturbations but $\sim100\%$ attack success rate) crafted from complex datasets like ImageNet. 
Our solution demonstrates great potentials to safeguard the DNNs against adversarial attacks in practical applications, given that it is likely the attackers prefer to generate stronger adversarial examples with minimum adversarial perturbations from realistic large-scale dataset so as to evade any possible defense methods.

\vspace{-8pt}
\subsubsection{White Box Mitigation}
% \vspace{-8pt}
\label{white}
In this section, we evaluate our method against recent BPDA attack, of which adversary knows the defense method and iteratively generates adversarial examples according to the defense. We implement our defense--Feature Distillation (FD-1P) in the released BPDA attack~\cite{athalye2018obfuscated} code at GitHub, using the same “Inception v3” model and 100 iterations for BPDA. The accuracy of benign examples (adversarial examples) after defense-$acc_l$ ($acc_m$), for different methods are reported in Table~\ref{t2}.

\begin{figure}[b]
\vspace{-12pt}
\centering
\includegraphics[width=0.75\columnwidth]{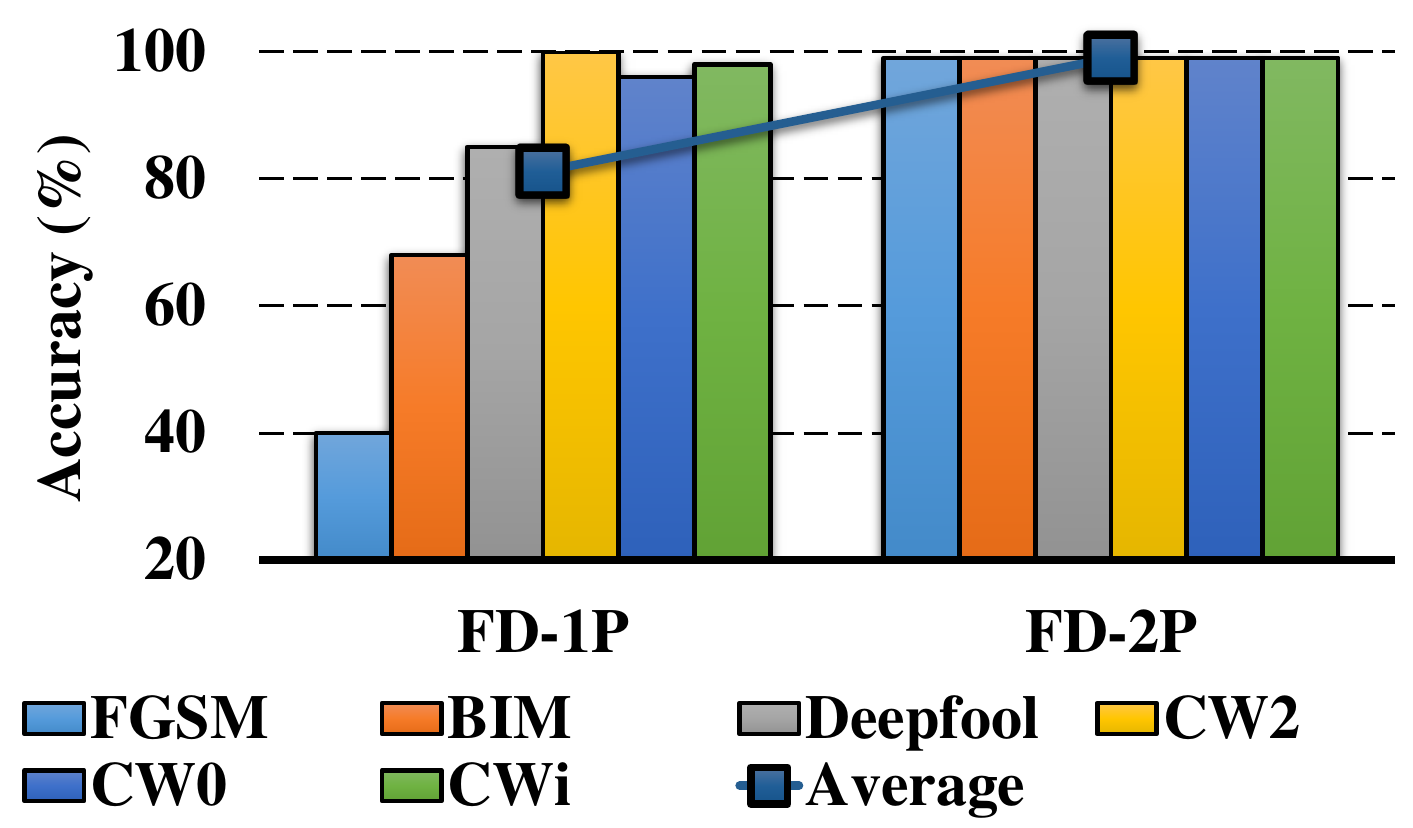}
\caption{\small Defense efficiency of black-box setting for different attack and defense mechanisms on ImageNet.}
\label{blackres}
\vspace{-10pt}
\end{figure}

%This results shown the 
\textbf{First}, Bit-depth, quilting and TVM does not offer any defense against BPDA, as expected. \textbf{Second}, JPEG can slightly mitigate BPDA by degrading image quality, i.e. quality factor from 75 to 10, defense efficiency ($acc_m$) is improved from 0 to 45\%. This is consistent with the recent result~\cite{shaham2018defending}.
%The authors also claim that "JPEG appears to be the most successful defense among all tested defenses, under FGA and BPDA.”
However, $acc_l$ drops by ~17\% compared to baseline (61\% v.s. 78\%), which is unacceptable. This reason is because  %quantization steps in JPEG increase as the frequency becomes high, 
in order to eliminate a large perturbation of BPDA attack in the lowest frequency component in JPEG, a significant large quantization factor (QF) will be needed. As a result, large quantization errors will occur in high frequency components, thereby significantly hurting $acc_l$.  
%which will incur large quantization errors or even zeroout in high frequency components,
%Quantization steps in JPEG have a increasing trend, in order to quantize a large perturbation in lowest frequency component large QF is required which will add large quantization error or directly zeroing-out higher frequency information, thus significantly hurting $acc_l$.
\textbf{Third}, On the other hand, our solution can provide the best defense efficiency against BPDA with least $acc_l$ reduction among all solutions, i.e. from FD (1X) to FD (3X), $acc_m$ is improved from 10\% to 60\%, with merely 1\%-3\% $acc_l$ reduction compared to original 78\%. FD-1$\times$, 2$\times$, 3$\times$ represent the quantization step (QS) of FD adopted in Table.~\ref{t1} (reference), 2 times and 3 times of the referred QS, respectively. 
%Note that our method outperform JPEG at both $acc_l$ and $acc_m$ 
%Quantization steps in JPEG have a increasing trend, in order to quantize a large perturbation in lowest frequency component large QF is required which will add large quantization error or directly zeroing-out higher frequency information, thus significantly hurting $acc_l$.
%Our method shown better defense efficiency and legitimate image testing accuracy than default JPEG is because \textcolor{red}{add }
\vspace{-8pt}
\subsubsection{Black Box Mitigation}
\label{black}
% \vspace{-8pt}
We follow the work~\cite{guo2017countering} for black-box analysis: DNN model used for testing is trained on transformed dataset (Feature Distillation), while attackers generate adversarial examples from the model trained on the original dataset. The crafted examples have high transferability between the two models for fair black-box analysis. We adopt “MobileNet” and the results of our methods are shown in Fig.~\ref{blackres}.

The average defense efficiency is improved from 56\%/91\% (Table~\ref{t1}) to 81\%/99\% (black-box) for our FD-1P/FD-2P method, respectively. These results indicate that our method defends against black-box setting efficiently. This is also consistent with the following conclusion based on~\cite{athalye2018obfuscated,guo2017countering}: Black box setting shows weak attack efficiency against the input-transformation based defenses.

% \vspace{-10pt}
\section{Conclusion}
% \vspace{-8pt}
As the robustness of DNN is significantly challenged by a variety of adversarial attacks, existing studies investigate the standard JPEG compression as a defense method, however, it is far from satisfactory in terms of both defense efficiency and testing accuracy. 
In this work, we propose the DNN-favorable feature distillation method by re-architecting the JPEG compression framework.
Compared with existing model-agnostic defense approaches, our ``feature distillation" can simultaneously reduce the adversarial attack success rate and maximize the testing accuracy on legitimate examples. Experimental results show that our method can improve the defense efficiency from $\sim20\%$ to $\sim90\%$ over most recent model-agnostic approaches with only marginal accuracy degradation ($\le1\%$), while significantly improving the processing time per image ($\sim260\times$ speedup). Our method also demonstrates the best defense efficiency against latest adaptive attack--BPDA ($\sim60\%$) with least accuracy drop ($\sim1\%$) when compared with other input-transformation based defenses.  
\section*{Acknowledgement}
This work is partially supported by the NSF under Grant CNS-1840813.

% \begin{table}[]
% \begin{tabularx}{\columnwidth}{llllllll}
%       & FGSM & BIM & DeepFool & CW2 & CW0 & CWi & Average \\
% FD-1P & 60   & 68  & 85       & 99  & 96  & 98  & 81.17   \\
% FD-2P & 99   & 99  & 99       & 99  & 99  & 99  & 99     
% \end{tabularx}
% \end{table}